\documentclass[11pt,a4paper]{article}
\usepackage[hyperref]{eacl2021}
\usepackage{times}
\usepackage{latexsym}

\usepackage{graphicx}
\usepackage{enumitem}
\usepackage{arabtex}
\usepackage{utf8}
\setcode{utf8}

\usepackage{microtype}

\aclfinalcopy 

\newcommand{\Bten}{QARiB10 }
\newcommand{\Btwentyfive}{QARiB25 }
\newcommand{\Btwentyfivemix}{QARiB25\_mix }
\newcommand{\Btwentyfivemixfar}{QARiB25\_mix\_far }
\newcommand{\Bsixtymix}{QARiB60\_mix }

\newcommand{\squeezeup}{\vspace{-5mm}}
\newcommand{\squeezeupless}{\vspace{-2mm}}

\title{Pre-Training BERT on Arabic Tweets: Practical Considerations}

\author{
Ahmed Abdelali ~~~ Sabit Hassan ~~~ Hamdy Mubarak
\\\bf{Kareem Darwish} ~~~ \bf{Younes Samih} \\
{\tt \{aabdelali,sahassan2,hmubarak,kdarwish,ysamih\}@hbku.edu.qa} \\ 
Qatar Computing Research Institute, HBKU Research Complex, Doha 5825, Qatar 
}

\date{}

\begin{document}
\maketitle
\begin{abstract}
Pretraining Bidirectional Encoder Representations from Transformers (BERT) for downstream NLP tasks is a non-trival task. We pretrained 5 BERT models that differ in the size of their training sets, mixture of formal and informal Arabic, and linguistic preprocessing.  All are intended to support Arabic dialects and social media.  The experiments highlight the centrality of data diversity and the efficacy of linguistically aware segmentation. They also highlight that more data or more training step do not necessitate better models. Our new models achieve new state-of-the-art results on several downstream tasks. The resulting models are released to the community under the name QARiB. 

\end{list}
\end{abstract}

\section{Introduction}

Recent advances in language modeling and word representations permitted breakthroughs in many natural language understanding (NLU) tasks, with both word level and sentence level tasks benefiting from such advancements. This includes static word embeddings (ex. word2vec, GloVe, etc.) and contextual embeddings (ex. BERT, RoBERTa, etc.) based on a transformer model. Contextual embeddings are able to capture much linguistic phenomena and properties, and when fine-tuned, they greatly improve the effectiveness of downstream tasks such named entity recognition (NER), text classification, etc. The effectiveness of contextual embeddings is enhanced when the domain of the text on which the embeddings were trained matches the domain of the text in downstream tasks, with more in-domain training data typically yielding improved results. 
In this paper we explore fundamental questions associated with training a transformer model from scratch including:  \textbf{i)} When is training a model from scratch warranted? \textbf{ii)} How much data is required to train an effective model?  \textbf{iii)} How many training iterations are required?  \textbf{iv)} and Is it better to use language-specific processing or SentencePiece (BP) segmentation? \\ 
We address these questions through extensive experimentation that we conducted in our efforts to train QCRI Arabic and Dialectal BERT (QARiB); an Arabic BERT model for downstream tasks involving tweets. While Modern Standard Arabic (MSA) is mostly used in formal communication, tweets are often dialectal, where dialects lack spelling conventions, informal in nature, and may include emojis, hashtags, user mentions. 
We trained 5 different transformer models of QARiB with data of different sources, sizes, and linguistic processing.  We compare the models against three existing BERT models, namely multilingual BERT (mBERT), which covers many languages including Arabic, AraBERT, which is an Arabic transformer model that was trained on a large MSA dataset and achieved improved results over mBERT on a variety of downstream tasks, and Arabic BERT (ArabicBERT), which is built by even larger collection from Arabic web-dump (OSCAR Corpus).  We aim to release 
10 checkpoints of QARiB, including the 5 best performing models. Potentially, this will help researchers speedup their development by providing a variety of training combinations.  The contributions of this paper are:
\begin{itemize}[leftmargin=*]
    \item We provide a detailed analysis to the above questions on the efficacy of training a transformer model and suggest best practices, based on training 5 QARiB models that cover different domains, data sizes, and linguistic processing.
    \item We compare our models to three other transformer models using a battery of tests.
    \item We release the pretrained models for the community to be used for fine tuning tasks~\footnote{Models available from https://huggingface.co/qarib}.
\end{itemize}



\section{Related Work}
The success of transformers model in many NLP tasks has motivated researchers to further deploy them in a range of downstream applications. This success was mainly motivated by the success in unsupervised fine-tuning approaches~\cite{Devlin_2019,radford2019language}. Though multilingual BERT trained on the Wikipedia articles from 104 languages facilitated the use of Transformer models for many models and the potential bridging between languages in a joint space, specialized monolingual BERT models routinely outperform multilingual BERT as seen for Chinese \citep{tian2020anchibert},  
French~\citep{Martin2020CamemBERTAT}, 
Dutch~\citep{delobelle2020robbert,devries2019bertje}, 
Finnish~\citep{virtanen2019multilingual}, Spanish~\citep{canete2020spanish}, and Turkish~\citep{stefan_schweter_2020_3770924} among others. 
For Arabic, \newcite{antoun2020arabert} pre-trained AraBERTv0.1 using 70 million sentences, about 24GB of text from news that covers different media outlets from various Arab regions. In another setting, \newcite{antoun2020arabert} preprocessed the text using the Arabic Farasa Segmenter~\cite{abdelali-etal-2016-farasa} to train AraBERTv1~\cite{djandji-etal-2020-multi} . These sources mostly use MSA. \newcite{safaya2020kuisail} trained Arabic-BERT on the Arabic portion of the OSCAR Corpus~\cite{ortiz-suarez-etal-2020-monolingual} and a Wikipedia dump. The total amount of data used sum up to 8.2 Billion words or 95GB of text. The latter contains some dialectal data. Both models were evaluated using a battery of benchmarks and outperformed Multilingual BERT (mBERT). \newcite{talafha2020multidialect} fine-tuned AraBERT  using 10M Arabic tweets from the NADI shared task. The additional tuning allowed them to achieve the highest score on NADI.  We compare such tuning to pre-training from scratch. 

\squeezeupless
\section{Data Collection}
\squeezeupless
\label{sec:length}
To maximize data coverage and generalization, we used large collections of both formal and informal texts.
The formal texts includes:
\begin{itemize}[leftmargin=*]
\item Arabic Gigaword Fourth Edition\footnote{LDC Catalogue LDC2009T30}: It contains 9 distinct newswire sources. 
The total data comprises of 2.7M documents and 32M sentences (1B words). 

\item Abu El-Khair Corpus~\cite{elKhair2016-1.5}:  It was scrapped from online Arabic newspaper websites from a wider number of Arab countries between Dec 2013 and Jun 2014.
The collection contains 58M sentences (1.5B words). 

\item Open Subtitles: \newcite{LISON16.947} compiled a collection of 2.6B sentences from a large database of movie and TV subtitles with more than 60 languages. We selected the Arabic part of the corpus, which contains 83.6M sentences (0.5B words). The data that is mostly conversational, and the text segments therein are typically shorter than sentences in news articles.  
\end{itemize}
For the informal text; the data was collected using the Twitter streaming API with the language filter set to Arabic (“lang:ar”). 
Our collection includes over 440M unique tweets (2.7B words) that were deduplicated from a tweet set collected between 2012 and 2020 and containing 1.4B tweets. To identify duplicate tweets for removal, we created normalized tweets as follows, we: removed diacritics, punctuation marks, and elongation letter (Kashida); normalized different forms of \textit{Alif} (\<أإآ>) to plain \textit{Alif} (\<ا>), \textit{Alif Maqsoura} (\<ى>) to dotted \textit{Ya} (\<ي>), and \textit{Taa Marbouta} (\<ة>) to Haa (\<ه>); 
converted Farsi and decorated letters to their corresponding Arabic letters (ex.: converting \<پ، گ> to \<ب، كـ> respectively); 
removed all URLs, user mentions, emojis, and numbers; replaced consecutive spaces and new lines with single spaces; and limited the number of times a character is repeated to a maximum of two repetitions (ex. ``cooooool'' $\rightarrow$ ``cool''). 

To prepare all the text for pre-training, we performed some pre-processing steps, namely we  tokenized the text using Farasa \cite{abdelali2016farasa} and removed diacritics and \textit{Kashida} (word elongation).  Additionally for tweets, we: normalized user mentions, URLs, and numbers to @USERNAME, URL, and NUM respectively; converted Farsi and decorated letters to corresponding Arabic letters; split hashtags; and restricted the number of times a letter was allowed to repeat to two repetitions.  We left emojis intact.  

We produced two versions of the corpus, where we performed word segmentation on all the text using Farasa \cite{abdelali2016farasa} for one, and not for the other.  Due to the agglutinative nature of Arabic, different clitics and morphemes compose one word, e.g. the word \<وكتابنا> 
(wktAbnA -- and our book) is composed of a prefix (w -- and), a stem (ktAb -- book), and a suffix (nA -- our). Further, for both versions, we employed language agnostic segmentation using Byte-Pair Encoding (BPE)~\cite{sennrich-etal-2016-neural}, instead of using WordPiece tokenization~\cite{wu2016googles}, which was used for the original BERT.  So for the Farasa segmented version, it was initially segmented using Farasa and subsequently segmented using BPE \textcolor{black}{which allowed us to limit the vocabulary size.}

 \squeezeupless
\section{Pretraining}
 \squeezeupless
For model pretraining, we used a Google Cloud TPU (v2). To opted for using a TPU, because we ran initial experiments on in-house Nvidia V100 GPU with 32G of VRAM and we estimated that training a single BERT model would take more than a year.  Using a TPU, training can be done in a few days. Using a TPU incurs the cost for using both Cloud Storage (GCS) and a Compute Engine (GCE). To reduce cost, we pre-processed and sharded the data on our local machines, and we used GCE for training on TPU only. 
We accessed the TPU via Google Colaboratory, or ``Colab''.  We used the GCE free tier, which has a maximum lifetime of 12 hours 
per session.

For our experiments, we choose several selections from our data. Table~\ref{data-stats} provides details for the data selected for each experiment. \Bten and \Btwentyfive are respectively composed of 40M and 100M random tweets. 
\Btwentyfivemix, and \Bsixtymix are composed of a balanced set of tweets and MSA data randomly selected from our data. \Btwentyfivemixfar is identical to \Btwentyfivemix except that it was segmented using Farasa. 

\begin{table}
\small
\centering
\begin{tabular}{lcrc}
\hline \textbf{Data} & \textbf{Size} & \textbf{Lines} & \textbf{Tokens} \\ \hline
~\Bten & 10GB & 40M & 1.1B \\
~\Btwentyfive  & 25GB & 100M & 2.8B  \\
~\Btwentyfivemix & 25GB & 122M & 2.7B \\
~\Btwentyfivemixfar & 25GB & 122M & 4.2B \\
~\Bsixtymix & 60GB & 330M & 7.6B \\
\hline
AraBERTv0.1 & 23GB & 70M & 3.0B \\
AraBERTv1 & 23GB & 70M & 3.0B \\
ArabicBERT & 95GB & - & 8.2B \\
mBERT & - & - & - \\
\hline
\end{tabular}
\caption{\label{data-stats} Experiments data statistics and additional models used for reference. 
}
\squeezeup
\end{table}
We used the original implementation of BERT\footnote{\url{https://github.com/google-research/bert/}} with default parameters.  
The pretraining was conducted on a single task, namely predicting random missing words with a masking probability of 15\%. Another common task is predicting the next sentence.  However, since tweet are short and commonly composed of single sentences, using next sentence prediction would not be suitable for tweets. When training using different datasets, we retained multiple checkpoints during training to observe the training progression and the efficacy of the models on the evaluation tasks at different points during training.  


  \squeezeupless
 \section{Evaluation}
  \squeezeupless
 \subsection{Evaluations Datasets}
  \squeezeupless
 \textbf{Named Entity Recognition}
 The training dataset was composed of a the entirety of the ANERCORP Arabic NER dataset \cite{benajiba2008arabic}, which is composed of roughly 150K tokens, and the tweet NER training set described by \newcite{darwish2014simple} composed of 24.5K tokens.  For the test set, we used the tweet test set of \newcite{darwish2013named} composed of 13.1K tokens. Tokens are annotated using IOB tags for \textit{persons}, \textit{organizations}, and \textit{locations}. \\
\textbf{Emotion detection} We used the SemEval 2018, Task 1 (subtask E-c) dataset \cite{mohammad-etal-2018-semeval}, which contains 4,381 tweets, split into  2,278, 585  and 1,518 for train, dev and test respectively. Each tweet is annotated for presence of one or more of 11 emotions, which cover Plutchik's \cite{plutchik1991emotions} 8 basic emotions in addition to \textit{love}, \textit{optimism}, and \textit{pessimism}.\\
\textbf{QADI Arabic Dialects Identification} The dataset~\cite{abdelali2020arabic} contains 540K traing tweets from 18 Arab countries with a test set of 3,303 test tweets ($\approx$ 184 tweets per country). \\ 
\textbf{Offensive language detection} We used the Arabic SemEval2020 Task 12 (OffensEval 2020) \cite{mubarak2020arabic,zampieri-etal-2020-semeval}. 
The training and test sets contain 8K and 2K tweets respectively, where tweets are tagged as \textit{offensive} or \textit{not offensive}.\\ 
\textbf{Sentiment Analysis} We used the AJGT corpus~\cite{Alomari10.1007/978-3-319-60042-0_66, Dahou10.1145/3314941} which is composed of 1,800 dialectal Jordanian and MSA tweets of which 900 are tagged as positive and 900 are tagged as negative.  The dataset has an 80/20 training/test split.  
 

\begin{figure}[pt]
\centering
\includegraphics[width=7cm,height=3cm,keepaspectratio]{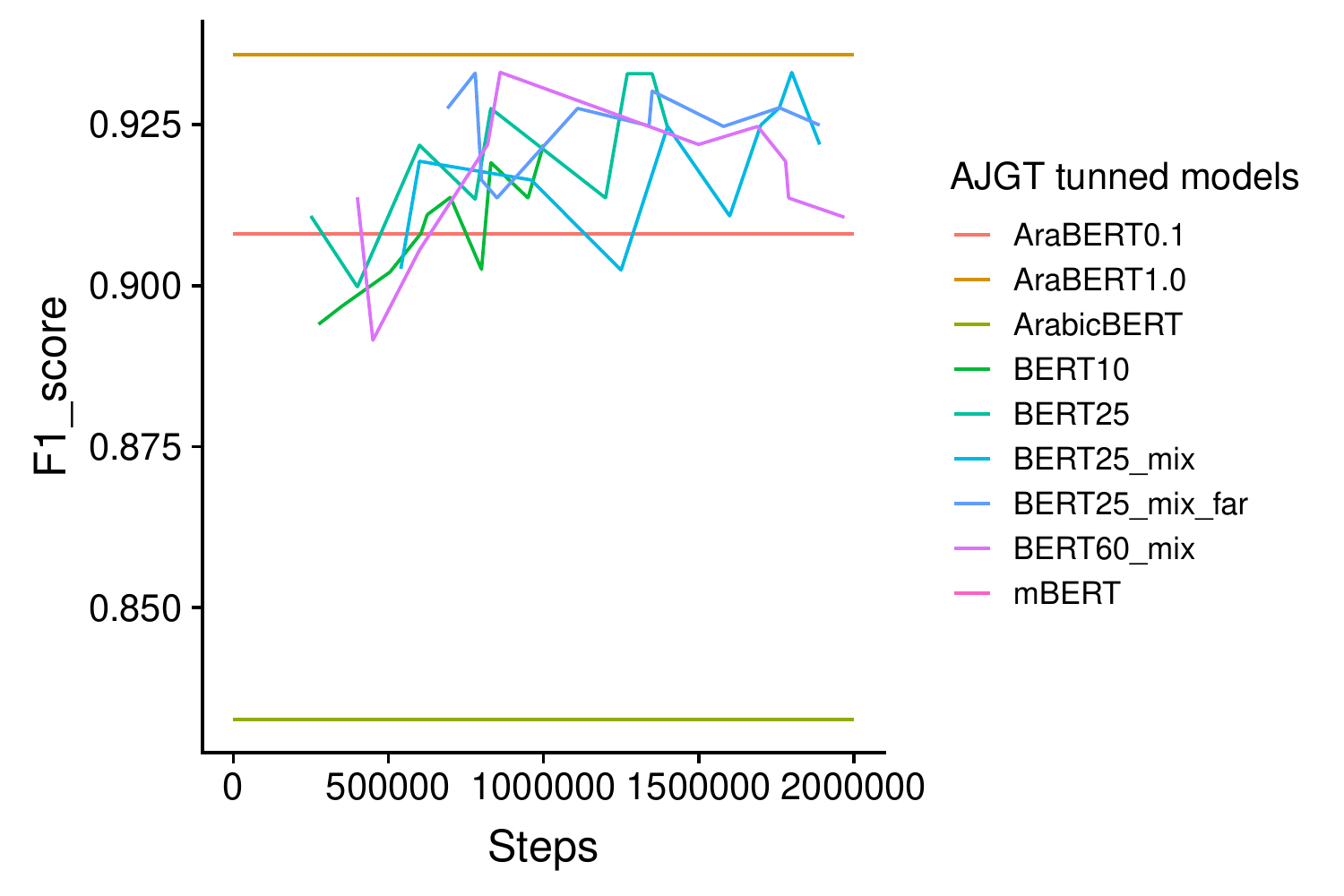} 
\caption{Results for fine tuning AJGT dataset}
\label{fig.AJGT}
\squeezeup
\end{figure}
 
\begin{figure}[h]
\centering
\includegraphics[width=6cm,height=3cm]{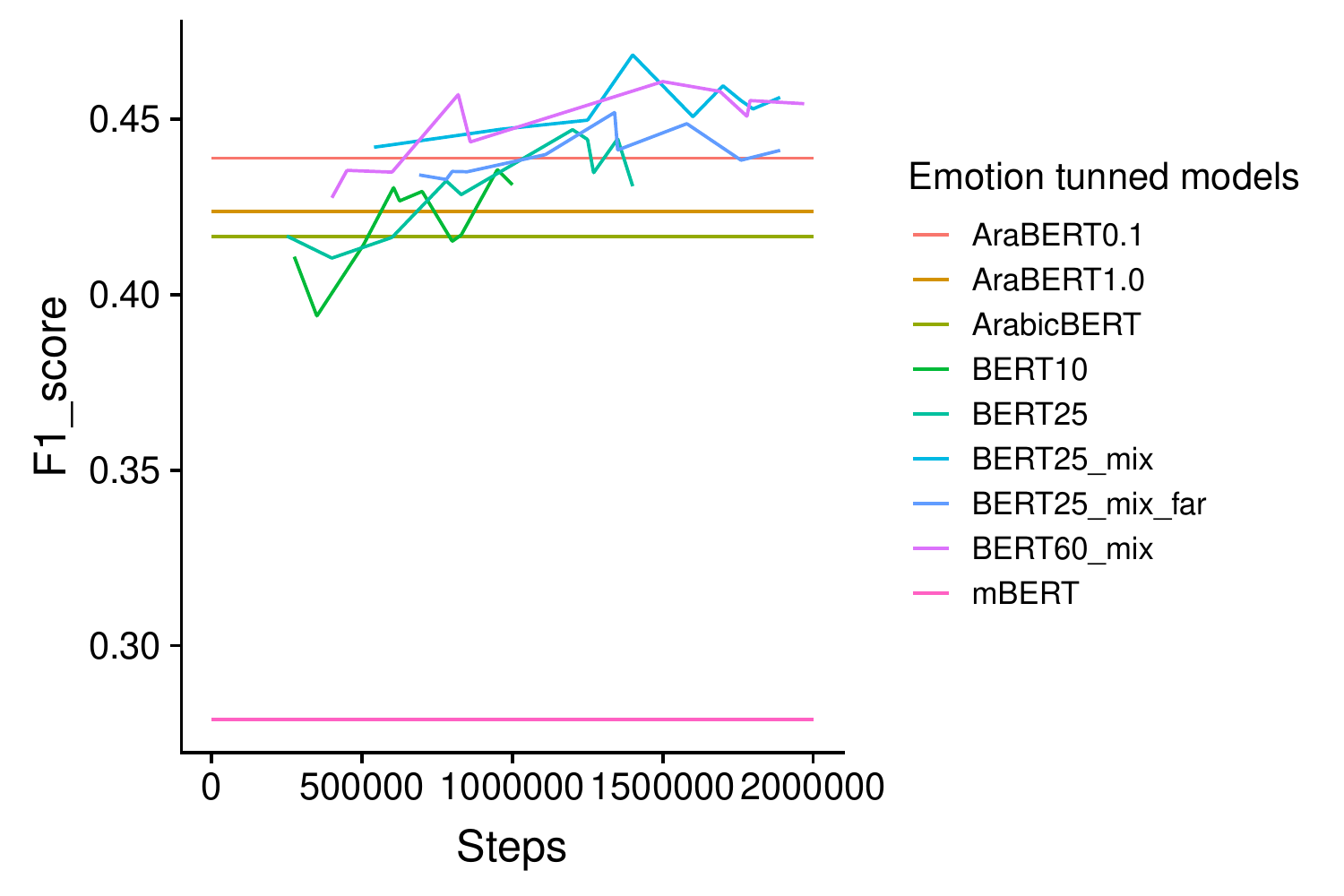} 
\caption{Results for fine tuning Emotion dataset }
\label{fig.emotion}
\squeezeup
\end{figure}

\begin{figure}[h]
\centering
\includegraphics[height=3cm]{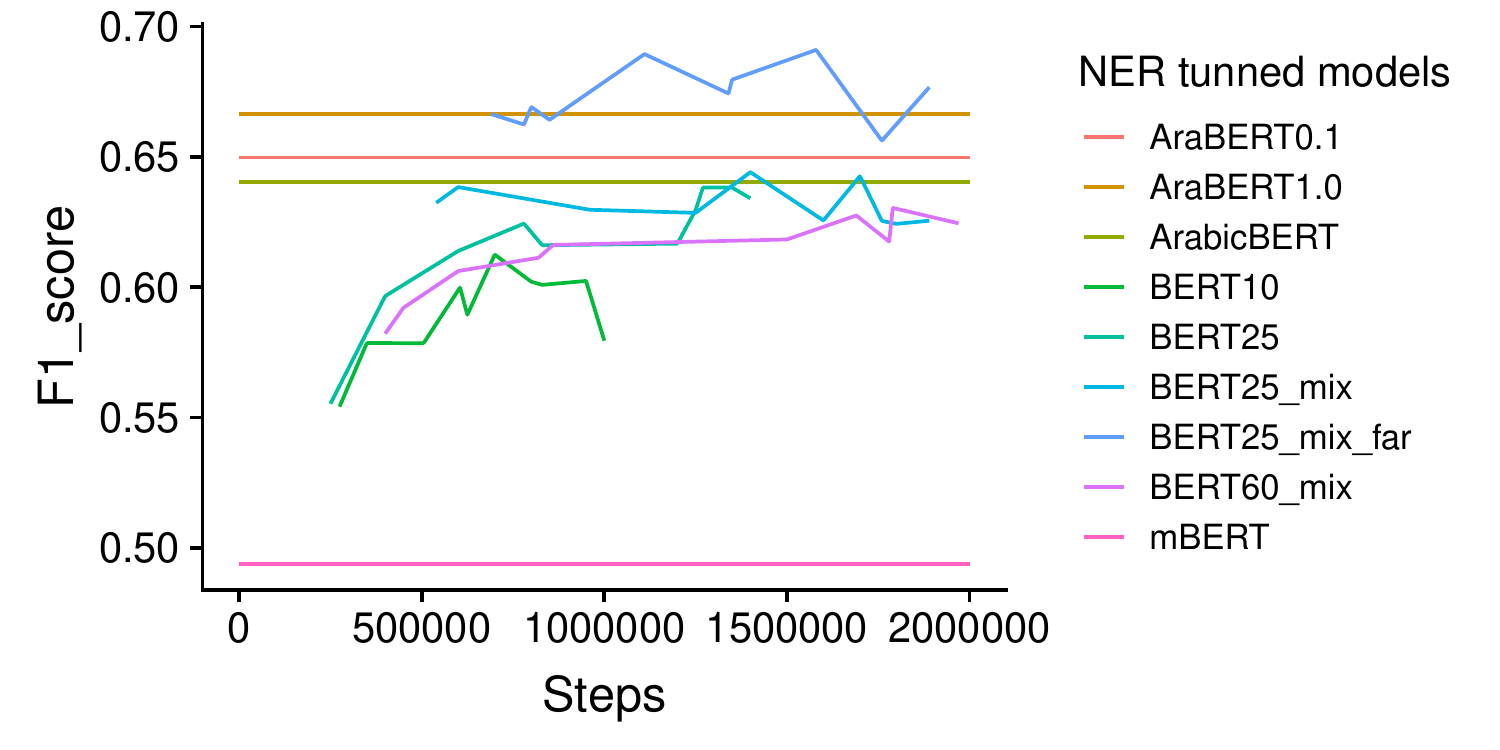} 
\caption{Results for fine tuning NER dataset}
\label{fig.NER}
\squeezeup
\end{figure}

\begin{figure}[h]
\centering
\includegraphics[width=6cm,height=3cm]{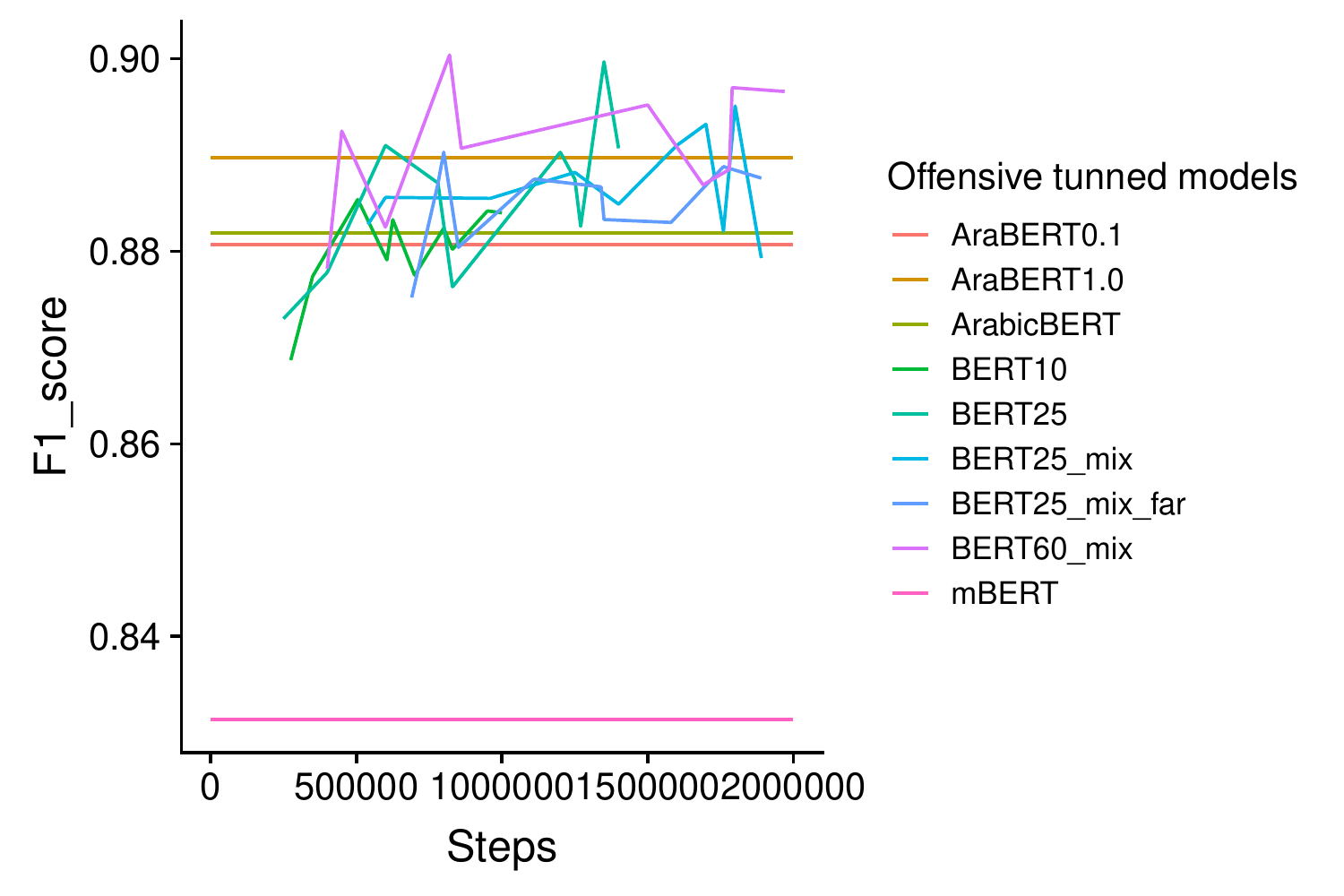} 
\caption{Results for fine tuning Offensive dataset }
\label{fig.Offensive}
\squeezeup
\end{figure}

\begin{figure}[h]
\centering
\includegraphics[width=5.5cm,height=3cm]{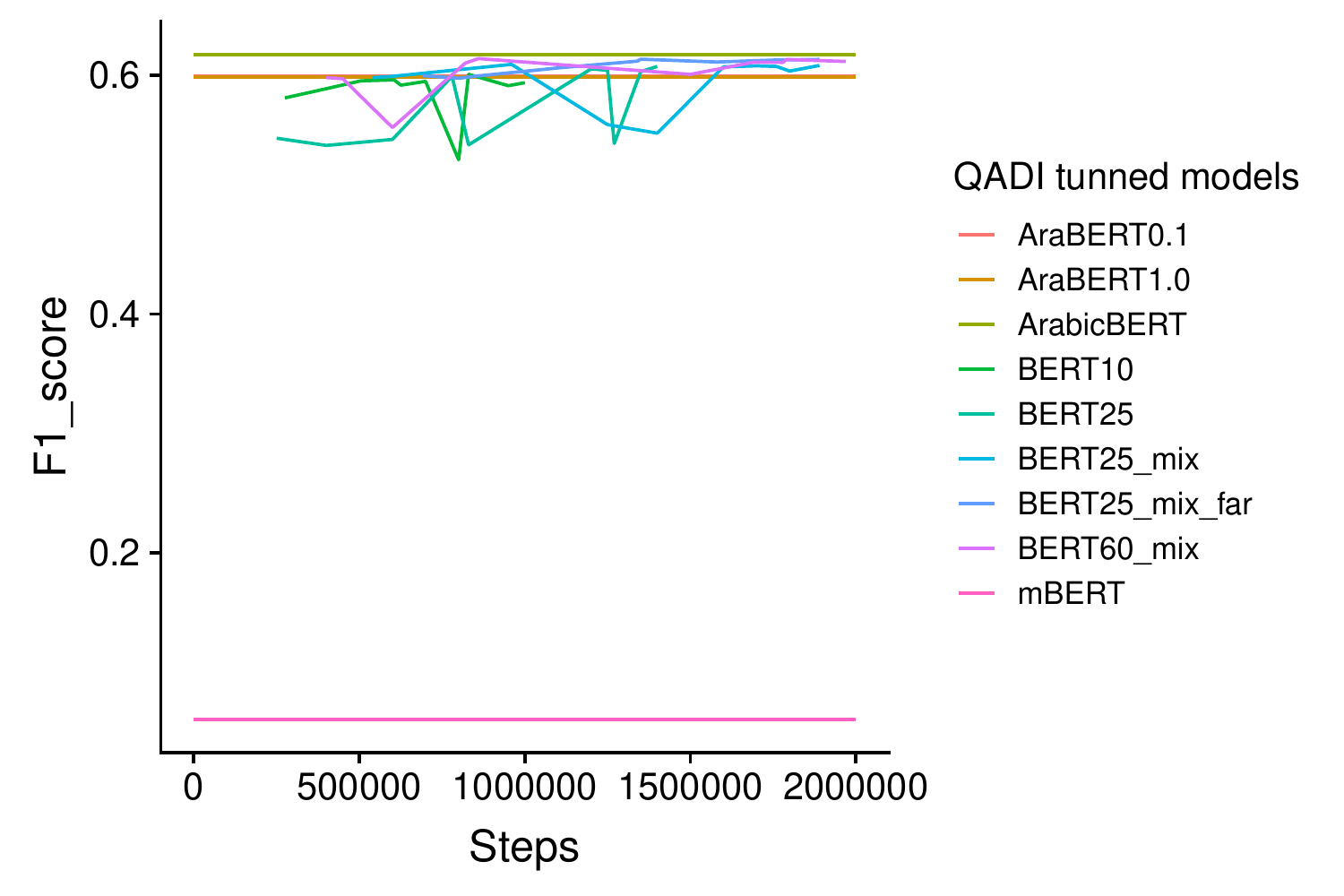} 
\caption{Results for fine tuning QADI dataset}
\label{fig.QADI}
\squeezeup
\end{figure}

\begin{table}
\centering
\small
\setlength\tabcolsep{2.5pt}
\begin{tabular}{@{}lccccc@{}}
\hline
Model & \textbf{AJGT} & \textbf{Emotion} & \textbf{NER} & \textbf{Offensive} & \textbf{QADI} \\ 
\hline
\Bten & 92.2 & 43.6 & 61.3 & 88.5 & 60.1  \\
\Btwentyfive & 93.3 & 44.7 & 63.8 & \textbf{90.0} & 60.7  \\
\Btwentyfivemix & 93.3 & \textbf{46.8} & 64.4 & 89.5 & 60.9  \\
\Btwentyfivemixfar & 93.3 & 45.2 & \textbf{69.1} & 89.0 & 61.3  \\
\Bsixtymix & 93.3 & 46.1 & 63.0 & \textbf{90.0} & \textbf{61.4}  \\
\hline 
AraBERTv0.1 & 90.8 & 43.9 & 65.0 & 88.1 & 59.9 \\
AraBERTv1 & \textbf{93.6} & 42.4 & 66.6 & 89.0 & 59.9  \\
ArabicBERT & 83.3 & 41.7 & 64.0 & 88.2 & \textbf{61.7}  \\
mBERT & 86.6 & 27.9 & 49.4 & 83.1 & 57.8  \\
\hline
\end{tabular}
\caption{\label{evaluation-results} Evaluation results}
\squeezeup
\end{table}

 \subsection{Evaluation Experiments} 
\squeezeupless
We compared performance of our QARiB 
models with AraBERTv0.1, AraBERTv1, ArabicBERT and mBERT. We used F1 score of minority class as our primary metric for binary classification and macro-averaged F1 for multi-class classification. For fair comparison, We compare the \textbf{base} (L=12, H=768, A=12, Total Parameters= 110M) settings from the different models. 
Figures ~\ref{fig.AJGT} to ~\ref{fig.QADI} show the assessment of various models at different checkpoints on the evaluation tasks. While typically the models perform higher using higher/more steps, this is not a monotonically increasing. It is noticeably common to observe fluctuations in effectiveness with more training steps. Thus, more training steps are not necessarily better and it is critical to continually test the model at different check points using multiple evaluation tasks to determine which checkpoint leads to the best results overall.  Further, perhaps using different checkpoints for a model could be used in combination in downstream tasks. Table~\ref{evaluation-results} summarizes the results of the different models using the checkpoint for each model that led the best overall results on the evaluation task and compares the results to other BERT models. From looking at the results, we can observe several things:  First, using more training data does not necessitate better models.  Improvements can only observed when going from \Bten to \Btwentyfive. Second, mixing both tweets and formal Arabic led to better results than using tweets only, even though the evaluation tasks that we tested on all involved tweets. Third, using linguistically motivated word segmentation led to substantial improvements for some tasks.  This is consistent with results observed for AraBERTv0.1 and AraBERTv1.0, where the latter employs Farasa for segmentation. 
\vspace{-6pt}
\section{Conclusion}
\vspace{-6pt}
We explored the impact of pre-training BERT from scratch on a MSA and dialectal data from Twitter. Our results show that: there is diminishing return for adding more data; using linguistic segmentation helps; improving the variety of training data with the inclusion of both formal and informal text is better than using informal text alone even when testing on informal text; relying on the loss to determine when to stop pre-training is misleading and it is better to test models at different checkpoints on a battery of tests to determine best models; pre-training from scratch may be warranted over fine-tuning an existing model on out of domain data, most likely due to mismatch in vocabulary.  
Lastly, pretraining BERT models requires considering many parameters, such as genre and data size. Additionally, having a battery of evaluation tasks ensures when to stop training and the selection of the best performing checkpoints. The impact of language dependent segmentation is clear. 
For future work, we would explore using dialectal segmenter as well as experimenting with other models of different sizes of training parameters.

\bibliography{eacl2021}
\bibliographystyle{acl_natbib}

\end{document}